\documentclass[lettersize,journal]{IEEEtran}
\usepackage{hyperref}
\usepackage{amsmath,amssymb,amsfonts,nccmath}
\usepackage{calrsfs}
\DeclareMathAlphabet{\pazocal}{OMS}{zplm}{m}{n}
\usepackage{mathtools} 
\usepackage{algorithmic}
\usepackage{algorithm}
\usepackage{graphicx}
\usepackage{textcomp}
\usepackage[dvipsnames,table,xcdraw]{xcolor}
\usepackage[utf8]{inputenc}
\usepackage{multirow}
\usepackage{textcomp}
\usepackage{longtable}
\usepackage{latexsym}
\usepackage{url}
\usepackage[switch]{lineno}
\usepackage[labelformat=parens]{subfig}
\usepackage[inline]{enumitem}
\usepackage{lipsum}
\usepackage{footnote}
\usepackage{comment}
\usepackage{caption}
\usepackage{cite}

\definecolor{darkblue}{RGB}{0, 0, 139} % Koyu mavi tanımı

\def\BibTeX{{\rm B\kern-.05em{\sc i\kern-.025em b}\kern-.08em
    T\kern-.1667em\lower.7ex\hbox{E}\kern-.125emX}}
\usepackage{balance}
\begin{document}

% \title{WBHT: Wasserstein Black Hole Transformer}
\title{WBHT: A Generative Attention Architecture for Detecting Black Hole Anomalies in Backbone Networks}

\author{%
  \IEEEauthorblockN{%
    Kiymet Kaya\IEEEauthorrefmark{1},\IEEEauthorrefmark{3},\IEEEauthorrefmark{5},
    Elif Ak\IEEEauthorrefmark{4}
    %Trung Q. Duong\IEEEauthorrefmark{4},
    Sule Gunduz Oguducu\IEEEauthorrefmark{2},\IEEEauthorrefmark{5},
  }% 
  \\
  \IEEEauthorblockA{\IEEEauthorrefmark{1}Istanbul Technical University, Department of Computer Engineering, Türkiye } \\
  \IEEEauthorblockA{\IEEEauthorrefmark{2}Istanbul Technical University, Department of Artificial Intelligence and Data Engineering, Türkiye }\\
\IEEEauthorblockA{
 \IEEEauthorrefmark{4}Memorial University, Canada} 
\IEEEauthorblockA{\IEEEauthorrefmark{3}BTS Group, Türkiye} \\
\IEEEauthorblockA{\IEEEauthorrefmark{5}ITU AI Research and 
  Application Center, Istanbul, Türkiye}

Email: {kayak16}@itu.edu.tr, elif.ak@mun.ca, sgunduz@itu.edu.tr}
%tduong@mun.ca, 

\markboth{IEEE International Symposium on Personal, Indoor and Mobile Radio Communications}%
{How to Use the IEEEtran \LaTeX \ Templates}

\maketitle
\begin{abstract}

We propose the Wasserstein Black Hole Transformer (WBHT) framework for detecting black hole (BH) anomalies in communication networks. These anomalies cause packet loss without failure notifications, disrupting connectivity and leading to financial losses. WBHT combines generative modeling, sequential learning, and attention mechanisms to improve BH anomaly detection. It integrates a Wasserstein generative adversarial network with attention mechanisms for stable training and accurate anomaly identification. The model uses long-short-term memory layers to capture long-term dependencies and convolutional layers for local temporal patterns. A latent space encoding mechanism helps distinguish abnormal network behavior. Tested on real-world network data, WBHT outperforms existing models, achieving significant improvements in F1 score (ranging from 1.65\% to 58.76\%). Its efficiency and ability to detect previously undetected anomalies make it a valuable tool for proactive network monitoring and security, especially in mission-critical networks.

\end{abstract}

%EA: conference'lara index term konmuyormus... :) 
%KK: aaaa :):) E bizim ICC'lerde var :D
\renewcommand\IEEEkeywordsname{Keywords}

% \cite{10622396}
\section{Introduction} \label{sec:intro}
The increasing reliance on communication networks for mission-critical applications, such as emergency services, industrial Internet of Things (IoT), autonomous transportation, and critical infrastructure monitoring, makes anomaly detection a paramount concern in ensuring both network security and reliability \cite{10612836}. In particular, anomalies like black holes (BH), which silently drop data packets without issuing failure notifications, pose significant threats to these mission-critical infrastructures \cite{10622177}. Due to their stealthy nature, they can cause prolonged undetected interruptions, severely impacting applications such as real-time monitoring in industrial IoT systems, disrupting command-and-control channels in autonomous transportation systems, and impeding timely communication in emergency response scenarios. 

% Existing machine learning (ML) and deep learning (DL)-based detection techniques, including supervised and unsupervised approaches, often struggle to detect these anomalies due to their stealthy nature and evolving patterns. Supervised methods require extensive labeled datasets, which are costly and time-consuming to curate, while unsupervised methods frequently suffer from high false alarm rates and limited ability to generalize to unseen anomalies.
Despite advancements in generative AI and sequential learning, existing solutions still face fundamental limitations in accurately identifying BH anomalies in backbone networks \cite{9852788}. Traditional autoencoder-based methods struggle to reconstruct complex network traffic, while adversarial learning techniques, such as generative adversarial networks (GANs), often suffer from unstable training and lack structured latent space representations for effective anomaly detection. Furthermore, existing transformer-based approaches, although promising in capturing long-range dependencies, fail to fully exploit spatial-temporal relationships within network traffic data. %As networks grow in complexity, driven by the increasing adoption of cloud computing, edge intelligence, and 6G architectures, the need for a robust and efficient BH anomaly detection framework remains an unsolved challenge.

%Despite Internet Control Message Protocol (ICMP), unexpected situations may occur in backbone networks, where routers drop data packets without sending any failure notifications and without notifying the sender, which adversely affects the end-to-end connection. In telecommunication networks, such faults are called ``black holes” or ``silent faults".  %A black hole (BH) may occur for a variety of reasons, from hardware malfunctions to misconfigurations and errors in individual router implementations, and for these reasons, it is very difficult to determine in advance. An error that occurs in the system is usually detected after a failure has occurred, which yields a large number of customers to be offline, or out of network, for extended periods. Prolonged communication disruption in the network and a complete loss of connection can cause significant financial losses to both the Internet Service Provider (ISP) and the customer. Therefore, from an operational point of view, it is imperative to design a mechanism that can predict the black hole before it occurs. 

%In this paper, we propose the Wasserstein Black Hole Transformer (WBHT) framework that combines the strengths of generative models, sequential learning, and attention mechanisms for BH anomaly detection in communication networks. 

Recognizing that each learning architecture possesses distinct advantages, it remains essential to empirically investigate their efficacy in specific anomaly detection contexts. Motivated by these challenges and building upon our prior research that employed an unsupervised convolutional autoencoder (Conv-AE) combined with density-based spatial clustering of applications with noise (DBSCAN) \cite{tnsm}, this study evaluates the performance of Wasserstein GANs integrated with attention mechanisms specifically for BH anomaly detection. While GANs and transformers have individually or jointly demonstrated robust performance across various anomaly scenarios, including intrusion detection, fraud detection, network traffic irregularities, sensor malfunctions, and industrial IoT anomalies \cite{9631286}, their combined capabilities have yet to be fully explored and validated in the critical case of BH anomalies. To validate the effectiveness of our WBHT, we utilize a substantial dataset in collaboration with an Internet technology provider \footnote{https://www.btsgrp.com}. The main contributions of this study can be summarized as follows:

%ma2024research
\begin{itemize}
% Transformer-based Sequential Anomaly Detection: 
    \item Unlike conventional GAN-based approaches, WBHT integrates transformer architectures to capture temporal dependencies in sequential network traffic.%, enhancing anomaly detection performance.

%Hybrid Wasserstein-GAN (WGAN) with Attention Mechanisms:
    \item WBHT leverages a hybrid approach using Wasserstein-GAN (WGAN) for stable training and improved anomaly score estimation, combined with multi-head attention to refine feature extraction from complex network data.
%Latent Space Encoding for Anomaly Characterization:     
%    \item Introduces an encoder-generator structure that maps network traffic into a well-structured latent space, improving the ability to reconstruct normal behavior and detect deviations indicative of BH anomalies.
    
%Scalable and Efficient Network Anomaly Detection:     
    \item WBHT benefits from transformer-based parallelization and Wasserstein loss stability, making it more efficient than traditional GAN-based detection methods, especially in large-scale network monitoring.
\end{itemize}

\begin{figure*}[htbp]
\centering
\includegraphics[width=.9\linewidth]{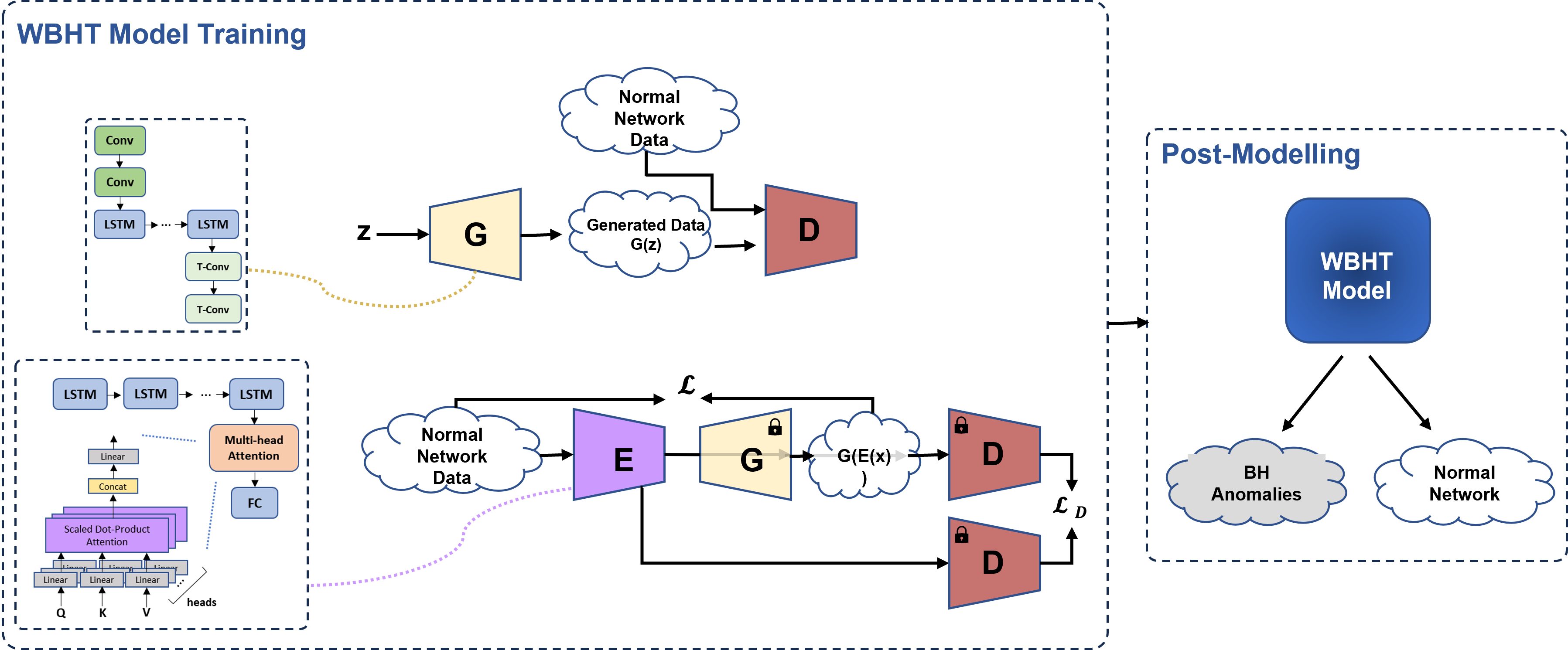}
%\vspace{3mm}
\caption{WBHT: Wasserstein Black Hole Transformer Anomaly Detection Framework.}
\label{fig_prop}
\end{figure*}
\vspace{-5pt}

\section{Literature Review} \label{sec:lite}
%Anomaly detection in computer communication is a well-established research area. 
Recent studies have concentrated on exploring unique characteristics of various anomaly types and leveraging available data through diverse machine learning approaches, ranging from supervised and semi-supervised to unsupervised techniques \cite{7723902}. %Supervised ML/DL approaches such as the ones predicting BH anomalies in the literature \cite{yasin2016feature, pandey2020blackhole, nagalakshmi2021machine, almomani2016wsn} rely on labeled datasets, aiming to fit a model that predicts how well new data fits. However, these models struggle with detecting previously unseen anomalies and face challenges due to the high cost and time involved in labeling data. To overcome this, unsupervised anomaly detection methods using unlabeled data have been proposed. These methods assume that most data points are normal, with only a small fraction being anomalous, and that normal data have a simple structure distinct from anomalies.

%Autoencoder (AE) models are particularly relevant here as they learn the underlying structure of data through unsupervised learning, identifying anomalies by detecting deviations from the learned structure. AE's ability to create synthetic data, aligned with the core concepts of generative AI, enhances the model's capacity to identify anomalies by generating examples that reflect the normal data distribution, further refining the detection process. Samuel et al. used a deep neural network model for network intrusion detection \cite{moore2023deep}. Convolutional Autoencoder (Conv-AE) \cite{khan2020toward} and Gated Recurrent Unit (GRU)-AE \cite{mushtaq2023knacks} models were also proposed for anomaly detection. The use of long-short-term memory AE (LSTM-AE) is suggested to detect anomalies in Internet routing \cite{muosa2022internet} and in wide area networks \cite{9262960}. 

Autoencoder (AE) models are particularly relevant for unsupervised anomaly detection, as they learn the underlying data structure and flag deviations as anomalies. Various AE architectures have been explored for this purpose, including deep neural networks for network intrusion detection \cite{moore2023deep}, convolutional autoencoders for capturing spatial patterns \cite{khan2020toward}, GRU-based AEs for sequential data \cite{mushtaq2023knacks}, and LSTM-AEs tailored to detecting anomalies in wide area networks \cite{9262960}.

While AE models are valuable in unsupervised intrusion detection, recent developments have led to the exploration of GANs, where they employ a generator and discriminator in an adversarial process. This setup enables GANs to detect anomalies by contrasting real and generated data, while also generating new samples that improve the model’s ability to distinguish normal from anomalous behavior. Among the notable approaches, DCT-GAN integrates dilated convolutions and Transformers within a GAN framework to improve time series anomaly detection by addressing mode collapse and enhancing generalization \cite{9626552}. TransEC-GAN combines a Transformer-enhanced GAN with Wasserstein distance and adaptive differential privacy for robust anomaly detection in industrial CPS \cite{10978569}.

Transformers, meanwhile, have advanced anomaly detection by effectively capturing temporal dependencies, outperforming traditional RNNs in sequential tasks. Adformer applies a Transformer-based adversarial framework to IoT sensor data, enhancing sensitivity to subtle anomalies through attention mechanisms \cite{Zeng}. Similarly, Transformer-based GANs have demonstrated superior capability over Autoencoder-based variants in generating high-quality adversarial samples for intrusion and malware detection, underscoring the advantage of attention-driven models in security contexts \cite{Laudanna}.

GANs combined with Transformer models have also been applied beyond time series and IoT contexts. One approach employs a Transformer-based GAN with Wasserstein GAN-GP for adversarial USB keystroke attack detection, improving robustness and accuracy \cite{Chillara}. Another framework leverages Transformer classifiers trained on GAN-generated data for intrusion detection, achieving state-of-the-art accuracy and enhanced resilience against evolving threats in metaverse security contexts \cite{Djenouri}.

Despite these advancements, challenges remain in detecting BH anomalies due to the difficulty of obtaining labeled datasets \cite{yasin2016feature, pandey2020blackhole, nagalakshmi2021machine, almomani2016wsn}. As a result, most studies rely on unsupervised or semi-supervised methods with limited or unlabeled data \cite{7723902}. In this study, we take a different approach by having access to a dataset known to contain exclusively normal traffic and use Wasserstein GANs with multi-head self-attention to learn its representation. This allows the model to detect anomalies as deviations without needing explicit BH examples, providing a practical and effective alternative to fully unsupervised approaches.

%A GAN-based framework is proposed for multivariate time-series anomaly prediction in cloud data centers, handling complex and dynamic data \cite{qi2024efficient}. GAN-based anomaly detection models are proposed for attack generation \cite{lin2022idsgan} and enhancing intrusion detection systems\cite{park2022enhanced,qi2023effective,li2019mad}.

%Diverging from traditional GAN-based methods, WBHT integrates transformers for capturing long-range dependencies and WGAN for stable training, making it more scalable and accurate in detecting BH anomalies in network traffic. Additionally, its encoder-generator structure enables a well-structured latent space representation, improving the model's ability to reconstruct normal behavior and detect deviations more effectively.

\section{Methodology} \label{sec:meth}

The flow of the proposed WBHT is given in Fig. \ref{fig_prop}. The model leverages a combination of generative modeling, sequential and adversarial learning, and attention mechanisms to accurately differentiate between normal network traffic and anomalous patterns indicative of BH anomalies. The training phase of WBHT consists of two primary components: the generative phase and the encoder-decoder phase.

The generative phase is based on the WGAN architecture, which improves training stability and alleviates common issues, such as mode collapse, often encountered in vanilla GANs. When comparing loss functions, WGAN uses the Wasserstein distance (WD) (also known as Earth Mover's Distance) as part of the loss function to learn the probability distribution, whereas GAN uses Jensen-Shannon (JS). JS takes values from 0 to $\log 2$, whereas WD addresses the issue that when JS distributions do not overlap, the derivative becomes 0 in the region where $\log 2$ applies. WGAN's WD is achieved by enforcing the Lipschitz constraint on the discriminator through weight-clipping\cite{arjovsky2017wasserstein}.

In the generative phase of WBHT, minimizing WD ensures a smooth and more meaningful optimization landscape, allowing the generator $G$ to produce high-quality network traffic representations that align with the normal data distribution. On the other hand, the discriminator $D$ is tasked with distinguishing between the real and generated data, optimizing this process through the WD, and ensuring stable and efficient adversarial learning. Here, $G$ combines LSTM layers for capturing long-term dependencies and convolutional layers for extracting local temporal patterns. The convolutional layer $C_{e}^{i}$ is formulated as in Equation \ref{eq_conv}, where \( W_i \) and \( b_i \) represent learnable weights and biases, while ReLU introduces non-linearity. The deconvolution layer $T_{d}^{i}$ is formulated as in Equation \ref{eq_t_conv}, focusing on reconstructing realistic sequences, where \( W_i^{'} \) and \( b_i^{'} \) denote transposed convolution filters and biases, ensuring temporal consistency in the generated data.

\begin{equation}
C_{e}^{i} (x_{i})= \text{ReLU}(W_{i}*x_{i} + b_{i}), \quad i=0,1,..,N
\label{eq_conv}
\end{equation}  

\begin{equation}
T_{d}^{i} (y_{i})= \text{ReLU}(W_{i}^{'} \otimes y_{i} + b_{i}^{'}), \quad i=0,1,..,N
\label{eq_t_conv}
\end{equation}  

WGAN training yields a generator \( G(z) \) that maps from the latent space \( Z \) (noise) to the data space \( X \) (training data), but does not provide the inverse mapping from \( X \) to \( Z \), which is essential for anomaly detection in WBHT. To address this limitation, an encoder $E$ is introduced, which transforms input network traffic into a compact latent representation. The encoder is trained separately, while keeping the parameters of the pre-trained $G$ and $D$ models fixed, such that it learns the inverse mapping \( x \to z \).

The encoder consists of stacked LSTM layers and a multi-head self-attention (MHSA) \cite{rogozhnikov2022einops} mechanism. The LSTM layers capture sequential dependencies, while MHSA enhances the model’s ability to focus on relevant time steps. The MHSA module processes three key components: the Query ($Q$), Key ($K$), and Value ($V$) vectors. The $Q$ computes similarity scores between the current time step and all other time steps, while the $K$ provides information about the time steps being compared. The $V$ contains the actual data at each time step. Attention scores are calculated using a dot product similarity function, determining the importance of each time step relative to the current one. These scores are then used to compute a weighted sum of the $V$ vectors, producing a context vector that retains critical information from the entire time series.

The overall loss function for WBHT model optimization is defined as in Equation \ref{eq1}. The first part of the Loss function comes from the Generator, and the second one is related to the Discriminator. Here, $f(\cdot)$ represents the discriminator features extracted from an intermediate layer, $n_d$ is the dimensionality of the intermediate feature representation, and $k$ is a weighting factor balancing the feature residual, and
$n$ denotes the total number of time steps in the input sequence.

%EA: taşmadı sanki böyle? oldu mu sence?
%KK: Bunu nasıl verelim bilemedim, böyle de satırdan taşıyor.
%%    \resizebox{.9\linewidth}{!}{    
\begin{equation}   
    \pazocal{L} = \frac{1}{n}\sqrt{\sum_{t=1}^{n} {x- G(E(x))}} + \\ \frac{k}{{n}_d}\sqrt{\sum_{t=1}^{n} {f(x)- f(G(E(x)))}}
    \label{eq1}
\end{equation}

The encoder-decoder phase refines the model’s ability to represent and reconstruct network traffic. The encoder $E$ transforms input network data into a lower-dimensional latent representation, which $G$ then attempts to reconstruct. The $D$ evaluates the reconstructed data, enhancing the model’s ability to detect deviations from learned normal patterns. This dual-stage training process enables WBHT to develop a deep understanding of the statistical characteristics of network traffic, making it robust against subtle and sophisticated BH anomalies.

After training, the WBHT model is deployed for post-modeling anomaly detection. Incoming network traffic is processed through the trained model, which classifies sequences as either normal network activity or BH anomalies. Classification relies on reconstruction errors and $D$’s confidence score. Sequences that significantly deviate from learned normal patterns are flagged as anomalies, enabling the identification of potential BH attacks in communication networks.

\begin{table*}[!htbp]
\caption{WBHT Performance: Evaluating GAN vs. WGAN with Different $E$ and $G$ Architectures}
\renewcommand{\arraystretch}{1.1}
\label{tab:res_gans}
\centering
\resizebox{.9\linewidth}{!}{
\begin{tabular}{lcccccccccccc}
\multicolumn{13}{c}{\large \color{RoyalBlue}\textbf{WGAN}} \\ \cline{2-13} 
 & \multicolumn{4}{c}{\textbf{G: FCNN}} & \multicolumn{4}{c}{\textbf{G: Conv.}} & \multicolumn{4}{c}{\textbf{G: LSTM}} \\ \cline{2-13} 
 & \textbf{DR} & \textbf{FAR} & \textbf{F1} & \textbf{Acc.} & \textbf{DR} & \textbf{FAR} & \textbf{F1} & \textbf{Acc.} & \textbf{DR} & \textbf{FAR} & \textbf{F1} & \textbf{Acc.} \\ \hline
E: FCNN & 0.9447 & 0.0842 & \cellcolor[HTML]{FDCF67}0.9174 & 0.9253 & 0.9430 & 0.0804 & \cellcolor[HTML]{FED467}0.9194 & 0.9272 & 0.9490 & 0.0825 & \cellcolor[HTML]{F9D567}0.9202 & 0.9278 \\ \hline
E: Conv & 0.9481 & 0.0825 & \cellcolor[HTML]{FED567}0.9199 & 0.9275 & 0.9498 & 0.0817 & \cellcolor[HTML]{E0D16C}0.9211 & 0.9286 & 0.9124 & 0.0747 & \cellcolor[HTML]{F9C269}0.9118 & 0.9211 \\ \hline
E: LSTM & 0.9498 & 0.0842 & \cellcolor[HTML]{FED467}0.9194 & 0.9269 & 0.9473 & 0.0870 & \cellcolor[HTML]{FCCD68}0.9164 & 0.9242 & 0.9354 & 0.0895 & \cellcolor[HTML]{F8BE6A}0.9101 & 0.9186 \\ \hline
E: ConvLSTM & 0.9473 & 0.0813 & \cellcolor[HTML]{F0D469}0.9205 & 0.9281 & 0.9379 & 0.0903 & \cellcolor[HTML]{F8BF6A}0.9105 & 0.9189 & 0.9179 & 0.0527 & \cellcolor[HTML]{FED567}0.9199 & 0.9275 \\ \hline
E: ConvMultiHead & 0.9515 & 0.0813 & \cellcolor[HTML]{C5CD72}0.9221 & 0.9294 & 0.9379 & 0.0767 & \cellcolor[HTML]{FBD666}0.9201 & 0.9281 & 0.9498 & 0.0796 & \cellcolor[HTML]{B7CB75}0.9226 & 0.9300 \\ \hline
E: LSTMMultiHead & 0.9507 & 0.0796 & \cellcolor[HTML]{AFCA77}0.9229 & 0.9303 & 0.9405 & 0.0796 & \cellcolor[HTML]{FED367}0.9190 & 0.9269 & 0.9481 & 0.0804 & \cellcolor[HTML]{D8D06E}0.9214 & 0.9289 \\ \hline
\multicolumn{13}{l}{} \\ \cline{2-13} 
 & \multicolumn{4}{c}{\textbf{G: ConvLSTM}} & \multicolumn{4}{c}{\textbf{G: ConvMultiHead}} & \multicolumn{4}{c}{\textbf{G: LSTMMultiHead}} \\ \cline{2-13} 
 & \textbf{DR} & \textbf{FAR} & \textbf{F1} & \textbf{Acc.} & \textbf{DR} & \textbf{FAR} & \textbf{F1} & \textbf{Acc.} & \textbf{DR} & \textbf{FAR} & \textbf{F1} & \textbf{Acc.} \\ \hline
E: FCNN & 0.9252 & 0.0957 & \cellcolor[HTML]{F2AA6D}0.9019 & 0.9111 & 0.9456 & 0.0800 & \cellcolor[HTML]{EBD36A}0.9207 & 0.9283 & 0.9311 & 0.0920 & \cellcolor[HTML]{F6B66B}0.9068 & 0.9156 \\ \hline
E: Conv & 0.9515 & 0.0829 & \cellcolor[HTML]{E6D26B}0.9209 & 0.9283 & 0.9371 & 0.0957 & \cellcolor[HTML]{F5B56B}0.9064 & 0.9150 & 0.9515 & 0.0821 & \cellcolor[HTML]{D5D06F}0.9215 & 0.9289 \\ \hline
E: LSTM & 0.9532 & 0.0792 & \cellcolor[HTML]{8BC47E}0.9242 & 0.9314 & 0.9456 & 0.0837 & \cellcolor[HTML]{FDD167}0.9181 & 0.9258 & 0.9558 & 0.0817 & \cellcolor[HTML]{A1C77A}0.9234 & 0.9306 \\ \hline
E: ConvLSTM & 0.9490 & 0.0767 & \cellcolor[HTML]{89C37F}0.9243 & 0.9317 & 0.9439 & 0.0862 & \cellcolor[HTML]{FCCB68}0.9157 & 0.9236 & 0.9532 & 0.0792 & \cellcolor[HTML]{8BC47E}0.9242 & 0.9314 \\ \hline
E: ConvMultiHead & 0.9515 & 0.0821 & \cellcolor[HTML]{D5D06F}0.9215 & 0.9289 & 0.9515 & 0.0817 & \cellcolor[HTML]{CDCE70}0.9218 & 0.9292 & 0.9524 & 0.0804 & \cellcolor[HTML]{ACC977}0.9230 & 0.9303 \\ \hline
E: LSTMMultiHead & 0.9575 & 0.0788 & \cellcolor[HTML]{57BB8A}0.9261 & 0.9331 & 0.9422 & 0.0846 & \cellcolor[HTML]{FCCC68}0.9162 & 0.9242 & 0.9532 & 0.0780 & \cellcolor[HTML]{76C083}0.9250 & 0.9322 \\ \hline
\multicolumn{13}{l}{} \\
\multicolumn{13}{c}{\large \color{RoyalBlue}\textbf{GAN}}  \\ \cline{2-13} 
 & \multicolumn{4}{c}{\textbf{G: FCNN}} & \multicolumn{4}{c}{\textbf{G: Conv.}} & \multicolumn{4}{c}{\textbf{G: LSTM}} \\ \cline{2-13} 
 & \textbf{DR} & \textbf{FAR} & \textbf{F1} & \textbf{Acc.} & \textbf{DR} & \textbf{FAR} & \textbf{F1} & \textbf{Acc.} & \textbf{DR} & \textbf{FAR} & \textbf{F1} & \textbf{Acc.} \\ \hline
E: FCNN & 0.8478 & 0.0802 & \cellcolor[HTML]{E67C73}0.8826 & 0.8964 & 0.9354 & 0.0858 & \cellcolor[HTML]{FAC469}0.9127 & 0.9211 & 0.9498 & 0.0837 & \cellcolor[HTML]{FED567}0.9197 & 0.9272 \\ \hline
E: Conv & 0.9184 & 0.0932 & \cellcolor[HTML]{F2A86D}0.9010 & 0.9106 & 0.9022 & 0.0982 & \cellcolor[HTML]{EB9070}0.8913 & 0.9019 & 0.9405 & 0.0870 & \cellcolor[HTML]{FAC769}0.9138 & 0.9219 \\ \hline
E: LSTM & 0.9396 & 0.0800 & \cellcolor[HTML]{FDD267}0.9184 & 0.9264 & 0.9507 & 0.0825 & \cellcolor[HTML]{E6D26B}0.9209 & 0.9283 & 0.9439 & 0.0854 & \cellcolor[HTML]{FCCC68}0.9162 & 0.9242 \\ \hline
E: ConvLSTM & 0.9379 & 0.0854 & \cellcolor[HTML]{FBC769}0.9140 & 0.9222 & 0.8963 & 0.1011 & \cellcolor[HTML]{E88672}0.8870 & 0.8981 & 0.9473 & 0.0813 & \cellcolor[HTML]{F0D469}0.9205 & 0.9281 \\ \hline
E: ConvMultiHead & 0.8997 & 0.1019 & \cellcolor[HTML]{E98872}0.8878 & 0.8986 & 0.9405 & 0.0809 & \cellcolor[HTML]{FDD167}0.9182 & 0.9261 & 0.9515 & 0.0825 & \cellcolor[HTML]{DDD16D}0.9212 & 0.9286 \\ \hline
E: LSTMMultiHead & 0.9388 & 0.0982 & \cellcolor[HTML]{F5B26C}0.9053 & 0.9139 & 0.8520 & 0.0759 & \cellcolor[HTML]{E98772}0.8872 & 0.9006 & 0.9507 & 0.0821 & \cellcolor[HTML]{DDD16D}0.9212 & 0.9286 \\ \hline
 & \multicolumn{1}{l}{} & \multicolumn{1}{l}{} & \multicolumn{1}{l}{} & \multicolumn{1}{l}{} & \multicolumn{1}{l}{} & \multicolumn{1}{l}{} & \multicolumn{1}{l}{} & \multicolumn{1}{l}{} & \multicolumn{1}{l}{} & \multicolumn{1}{l}{} & \multicolumn{1}{l}{} & \multicolumn{1}{l}{} \\ \cline{2-13} 
 & \multicolumn{4}{c}{\textbf{G: ConvLSTM}} & \multicolumn{4}{c}{\textbf{G: ConvMultiHead}} & \multicolumn{4}{c}{\textbf{G: LSTMMultiHead}} \\ \cline{2-13} 
 & \textbf{DR} & \textbf{FAR} & \textbf{F1} & \textbf{Acc.} & \textbf{DR} & \textbf{FAR} & \textbf{F1} & \textbf{Acc.} & \textbf{DR} & \textbf{FAR} & \textbf{F1} & \textbf{Acc.} \\ \hline
E: FCNN & 0.9303 & 0.0870 & \cellcolor[HTML]{F8BD6A}0.9099 & 0.9186 & 0.9490 & 0.1341 & \cellcolor[HTML]{E78073}0.8844 & 0.8931 & 0.9498 & 0.0833 & \cellcolor[HTML]{FED666}0.9200 & 0.9275 \\ \hline
E: Conv & 0.9362 & 0.0788 & \cellcolor[HTML]{FDD167}0.9180 & 0.9261 & 0.9515 & 0.0895 & \cellcolor[HTML]{FCCC68}0.9162 & 0.9239 & 0.9490 & 0.0796 & \cellcolor[HTML]{BFCC73}0.9223 & 0.9297 \\ \hline
E: LSTM & 0.9532 & 0.0825 & \cellcolor[HTML]{CDCE70}0.9218 & 0.9292 & 0.9464 & 0.0833 & \cellcolor[HTML]{FED267}0.9187 & 0.9264 & 0.9532 & 0.0804 & \cellcolor[HTML]{A4C879}0.9233 & 0.9306 \\ \hline
E: ConvLSTM & 0.9524 & 0.0813 & \cellcolor[HTML]{BDCC74}0.9224 & 0.9297 & 0.9507 & 0.0821 & \cellcolor[HTML]{DDD16D}0.9212 & 0.9286 & 0.9507 & 0.0784 & \cellcolor[HTML]{96C67C}0.9238 & 0.9311 \\ \hline
E: ConvMultiHead & 0.9481 & 0.0776 & \cellcolor[HTML]{A1C77A}0.9234 & 0.9308 & 0.9490 & 0.0813 & \cellcolor[HTML]{E0D16C}0.9211 & 0.9286 & 0.9473 & 0.0780 & \cellcolor[HTML]{B2CA76}0.9228 & 0.9303 \\ \hline
E: LSTMMultiHead & 0.9532 & 0.0796 & \cellcolor[HTML]{94C57D}0.9239 & 0.9311 & 0.9473 & 0.0821 & \cellcolor[HTML]{FED567}0.9199 & 0.9275 & 0.9515 & 0.0796 & \cellcolor[HTML]{A7C879}0.9232 & 0.9306 \\ \hline
\end{tabular}
}
\end{table*}

\section{Experimental Results} \label{sec:eval}
The real ISP network data containing BH traffic used in this study is collected, processed and cleaned according to procedures explained in our previous study \cite{tnsm}. % using the YANG\footnote{https://yangcatalog.org/} data modeling language with NETCONF protocol. More specifically, two YANG models, Cisco-IOS-XR-infra-statsd-oper and Cisco-IOS-XR-ip-rib-ipv4-oper are used \cite{10622396}. The obtained dataset is a time series data recorded in 5-minute periods from 01-07-2021 to 30-08-2021 and includes $17,280$ samples. To observe the performance of the BH prediction models, we divided the 60-day dataset into two, preserving the time order of samples in the dataset: the training set consists of $12,096$ instances (70\% of all the data) and test set consists of $5,184$ instances (30\% of all the data). In addition, $2,592$ samples in training data (15\% of all the data) were used as the validation set. 
%Next, we implemented several preprocessing steps to enhance the prediction capabilities of our proposed model. These steps included: (i) eliminating non-informative features, specifically those that maintained the same value across 90\% of the observations; (ii) identifying and removing redundant features by detecting pairs with a correlation value exceeding 0.9 and subsequently removing one from each pair; and (iii) generating new features through the modification and combined utilization of existing features. Moreover, features in the BH network traffic dataset are normalized to standardNorm, $({{x_{i}}^{j}}) = (x_{i}^{j}-\mu_{i})/\sigma_{i}$, where $x_{i}^{j}$ represents the $j$'th value of $i$'th feature, $\mu_{i}$ is the mean of the training samples, and $\sigma_{i}$ is the standard deviation of the training samples. The corresponding feature values in the test set are also normalized using the same $\mu_{i}$ and $\sigma_{i}$ values calculated on the training set.
Beyond the previous study, this approach gives a semi-supervised learning performed with all candidate forecasting models and the BH labeled samples were used only in the evaluation of the test set results. The subsections elaborate on the following aspects: Section \ref{sec:formationWBHT} presents the formation of the WBHT, while Section \ref{sec:eval-baselines} provides a comparative analysis of state-of-the-art models with WBHT for BH detection.

\subsection{Formation of the Wasserstein Black Hole Transformer}
\label{sec:formationWBHT}
We conducted a series of experiments to develop the WBHT model, as summarized in Table \ref{tab:res_gans}. Our primary objectives are: (i) to compare the performance of WGAN and vanilla GAN, (ii) to identify the most effective $E$ model, and (iii) to determine the best-performing $G$ model. These experiments select an optimal architecture that enhances model robustness and generalization.

We evaluated various $E$ and $G$ architectures, including Fully Connected Neural Networks (FCNN), Conv, LSTM, ConvLSTM, and Transformer-based architectures incorporating Multi-Head Attention mechanisms (ConvMultiHead, LSTMMultiHead). To assess model performance, we utilized several key evaluation metrics: Detection Rate (DR), False Alarm Rate (FAR), F1 Score (F1), and Accuracy (Acc). Among these, the F1 Score was prioritized for model selection due to its balanced consideration of precision and recall, which is crucial for mitigating the trade-off between false positives-negatives.

Based on the results, the WGAN consistently outperformed the vanilla GAN across multiple configurations. Additionally, among $E$ architectures, the LSTMMultiHead encoder demonstrated superior performance, particularly in capturing temporal dependencies and complex patterns. For $G$ architectures, the ConvLSTM model achieved the best results, effectively balancing spatial and temporal feature extraction. As a result of these experiments, WBHT was formed as a model incorporating WGAN as the generative AI method, LSTMMultiHead as the $E$, and ConvLSTM as the $G$. This combination leverages the strengths of both Transformer-based encoders and spatiotemporal feature extractors, ultimately improving overall model performance.

%KK: TimeSeriesTansforme-r r görünen versiyonuyla replace'licem.
%KK: Done!
\begin{figure*}
    \centering
    \includegraphics[width=.8\linewidth]{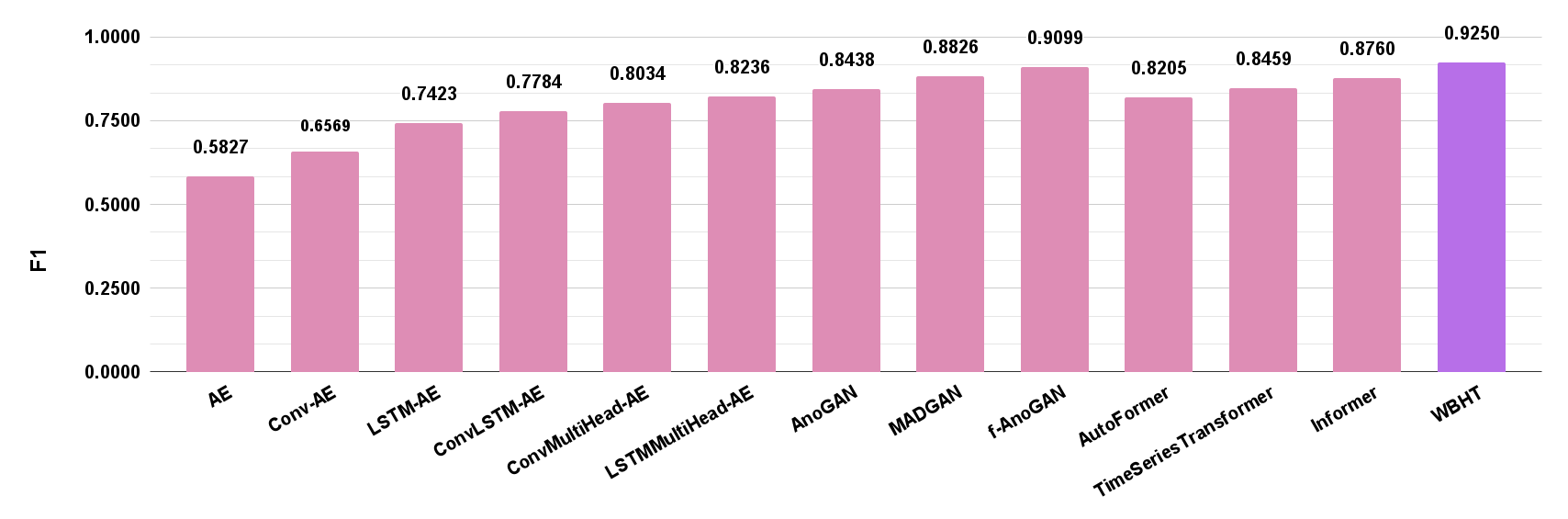}
    \includegraphics[width=.8\linewidth]{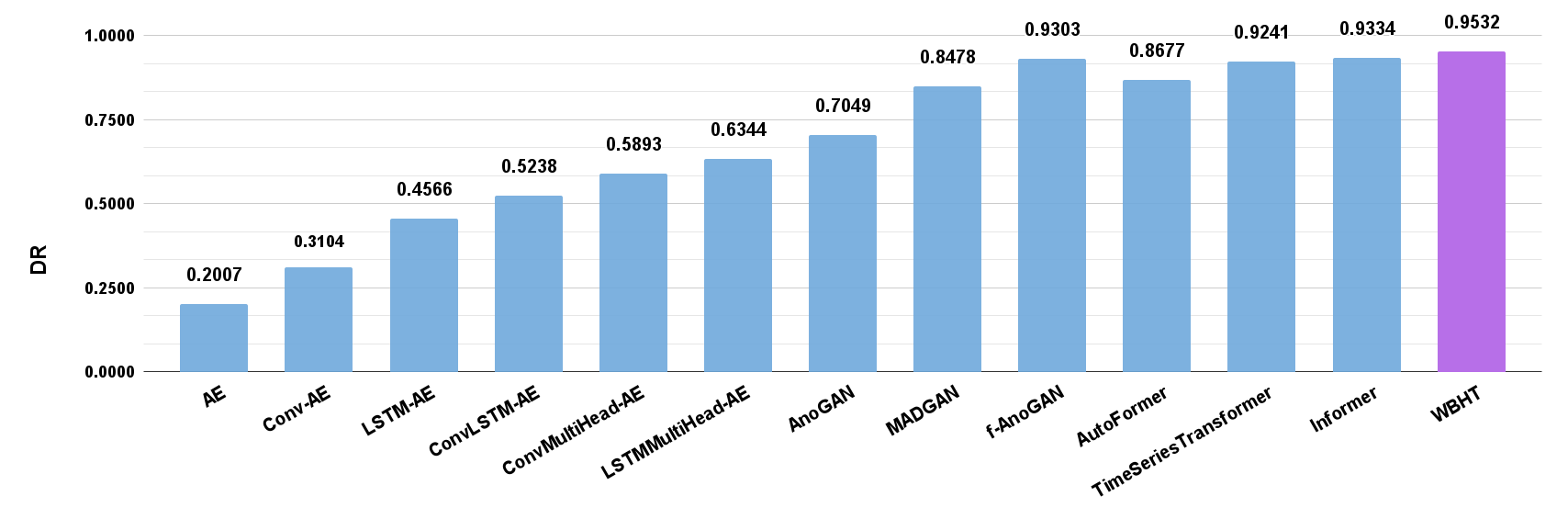}
    \caption{Comparative Performance Evaluation of BH Detection Models}
    \label{fig:comparative}
\end{figure*}
\vspace{-5pt}

\subsection{WBHT Black Hole Detection Evaluation} \label{sec:eval-baselines}
To validate the effectiveness of our proposed WBHT model, we compare its performance against state-of-the-art baseline models, as explained in the following:
% \cite{nixon2020autoencoders}

\textbf{AE} consists of only $Linear$ layers. This model has four $Linear$ layers, the first two used for encoding and the last two for decoding, which respectively transform \{\#features, 16, 8, 16, \#features\}. %After each $Linear$ layer, `ReLU' is used as an activation function.

\textbf{Con-AE} has 3 $Conv1D$ and 3 $T-Conv1D$ layers, the $filters$ are 32, 16, 8, 16, 32, 1 and $dropout$ $rates$: 0.2, 0, 0.2, 0, 0 from the left side, respectively.
%Other hyperparameter values are the same for all layers and $kernel size:$ 7, $padding:$ `same', $strides:2$, $activation$ (except last T-Conv1D layer): 'ReLU'. $Reconstruction$ $Loss:$ `$mean$ $absolute$ $error$ $(MAE)$' and model optimization is provided with the $Adam$ $algorithm$ and the $learning$ $rate$ is 0.001.
% \cite{longari2020cannolo}
\textbf{LSTM-AE} consists of $LSTM$ encoder and decoders with the hyperparameter values: $hidden size$: 8. 

\textbf{ConvLSTM-AE} is Con-AE's version with an $LSTM$ layer added between encoding and decoding.

\textbf{ConvMultiHead-AE} is Con-AE's version with MHSA mechanisms with $number of attention heads$: 4 after first $Conv1D$ and $T-Conv1D$. MHSA allows the model to attend to different parts of the time series simultaneously, capturing long-range dependencies and temporal patterns in the data.

\textbf{LSTMMultiHead-AE} is LSTM-AE's version with MHSA mechanisms with $number of attention heads$: 4 

%AnoGAN, MADGAN, f-AnoGAN kısalacak.
\textbf{AnoGAN \cite{anogan2017}} is an anomaly detection model based on vanilla GAN with a single $G$ and a single $D$. AnoGAN is trained using the JSD loss, which can sometimes cause unstable training. A major drawback of AnoGAN is the absence of an $E$, requiring an iterative optimization process to find the best latent representation of an input during inference. %This results in slow anomaly detection and makes the method computationally expensive. Additionally, since AnoGAN relies on a single discriminator, it is prone to mode collapse, where the generator fails to learn the full diversity of normal data, reducing its ability to detect anomalies effectively.
%The $G$ learns to generate normal samples, while the $D$ differentiates between real and synthetic samples.

%MADGAN (Multivariate Anomaly Detection GAN)
\textbf{MADGAN \cite{li2019mad}} improves time series anomaly detection by introducing multiple $D$'s but suffers from training instability due to its vanilla GAN's JSD and lacks an $E$ for fast inference like f-AnoGAN. However, since MADGAN still uses a vanilla GAN's JSD loss instead of WGAN-GP, training stability can be an issue, particularly in complex datasets. Additionally, MADGAN does not include an $E$, meaning it does not benefit from the fast inference provided by f-AnoGAN.

\textbf{f-AnoGAN \cite{schlegl2019f}} improves upon AnoGAN by introducing an $E$ network, which maps real input data to the latent space, bypassing the need for optimization, in addition to $G$ and $D$. f-AnoGAN also replaces JSD loss with WGAN-GP, resulting in more stable training and a better latent space representation.

\textbf{AutoFormer} \cite{wu2021autoformer} employs a decomposition-based architecture that separates time series into trend and seasonal components as $X_t = \text{AvgPool}(\text{Padding}(X)), \quad X_s = X - X_t$ where \( X_t \) is the smoothed trend component, and \( X_s \) captures seasonal variations. Additionally, Autoformer introduces an \textit{Auto-Correlation Mechanism} to identify periodic dependencies efficiently. %and by leveraging Fast Fourier Transform, it reduces complexity to \( O(L \log L) \), ensuring robust long-term forecasting.

\textbf{TimeSeriesTransformer} is an encoder-decoder architecture designed for long-term time series forecasting. It leverages self-attention to capture long-range dependencies efficiently, making it suitable for applications like energy forecasting, finance, and health monitoring. Its self-attention mechanism computes dependencies between inputs as $A(Q, K, V) = \text{Softmax} \left( (Q K^T) / \sqrt{d} \right) V$ where \( Q \), \( K \), and \( V \) represent query, key, and value matrices, and \( d \) is the dimensionality. 

\textbf{Informer} \cite{zhou2021informer} addresses the computational inefficiencies of self-attention by introducing \textit{ProbSparse Self-Attention}, which selectively focuses on the most significant attention scores. %This reduces memory and time complexity while preserving essential information. 
The attention mechanism is defined as $A(Q, K, V) = \text{Softmax} \left( (Q'K^T) / \sqrt{d} \right) V$, where \( Q' \) contains only the top-\( u \) queries based on sparsity constraints, reducing complexity. % to \( O(L \log L) \). 

\begin{table}[h]
\caption{Performance Benchmarking of WBHT}
\label{tab:baselinecomparison}
\centering
\setlength{\tabcolsep}{6pt}
\renewcommand{\arraystretch}{1}
\begin{tabular}{lcccc}
\hline
 & \textbf{DR} & \textbf{FAR} & \textbf{F1} & \textbf{Acc.} \\ \hline
\textbf{AE}                    & 0.2007 & 0.0062 & \cellcolor[HTML]{E67C73}0.5827 & 0.7347 \\
\textbf{Conv-AE}               & 0.3104 & 0.0136 & \cellcolor[HTML]{ED976F}0.6569 & 0.7656 \\
\textbf{LSTM-AE}               & 0.4566 & 0.0190 & \cellcolor[HTML]{F6B76B}0.7423 & 0.8097 \\
\textbf{ConvLSTM-AE}           & 0.5238 & 0.0194 & \cellcolor[HTML]{FAC569}0.7784 & 0.8314 \\
\textbf{ConvMultiHead-AE}      & 0.5893 & 0.0309 & \cellcolor[HTML]{FCCE68}0.8034 & 0.8450 \\
\textbf{LSTMMultiHead-AE}      & 0.6344 & 0.0330 & \cellcolor[HTML]{FFD666}0.8236 & 0.8583 \\
\textbf{AnoGAN}                & 0.7049 & 0.0499 & \cellcolor[HTML]{DED16D}0.8438 & 0.8700 \\
\textbf{MADGAN}                & 0.8478 & 0.0802 & \cellcolor[HTML]{9EC77A}0.8826 & 0.8964 \\
\textbf{f-AnoGAN}              & 0.9303 & 0.0870 & \cellcolor[HTML]{71C084}0.9099 & 0.9186 \\
\textbf{AutoFormer}            & 0.8677 & 0.2183 & \cellcolor[HTML]{FED467}0.8205 & 0.8394 \\
\textbf{TimeSeriesTransformer} & 0.9241 & 0.2485 & \cellcolor[HTML]{DBD16D}0.8459 & 0.8673 \\
\textbf{Informer}              & 0.9334 & 0.1922 & \cellcolor[HTML]{A9C978}0.8760 & 0.8920 \\
\textbf{WBHT (proposed)}       & \textbf{0.9532} & \textbf{0.0780} & \cellcolor[HTML]{57BB8A}\textbf{0.9250} & \textbf{0.9322} \\ \hline
\end{tabular}
\end{table}

%EA: su asagidaki yeri bir onceki cumleleri de koruyarak eklemeler yaptim ama bir bakar misin? bir sey silmek istemedim eksik bilgi olmasin diye ama fazlalik oldu gibi de bu sefer :) control edebilirsin summarize edek 
The results are presented in Table \ref{tab:baselinecomparison}. The complexity requirements of our model are assessed by benchmarking it against more primitive AE-based methods, namely LSTM-AE, ConvLSTM-AE. These models, while effective in capturing temporal dependencies, struggle with high-dimensional network traffic patterns and lack the adversarial learning necessary to distinguish subtle BH anomalies. Additionally, we investigate the adequacy of incorporating only Transformer-based architectures by comparing WBHT against Transformer-based baselines. %Specifically, we distinguish between primitive Transformer-based architectures, such as ConvLSTM-MultiHeadAE, and more advanced Transformer architectures, including Informer, AutoFormer, and TimeSeriesTransformer. 
While primitive Transformer-based models, such as ConvLSTM-MultiHeadAE, introduce self-attention mechanisms for improved feature extraction and temporal dependencies, they still lack the generative modeling needed for anomaly detection in general. More advanced Transformer architectures, including Informer, AutoFormer, and TimeSeriesTransformer, excel in capturing long-range dependencies in anomalies as an enhancement but fail to effectively localize anomalies. This is mainly due to the key characteristic of BH anomalies, occurring over short, bursty time intervals. Finally, we contrast our approach with advanced generative models that do not incorporate Transformers, such as AnoGAN, f-AnoGAN, and MADGAN. This comparison also enables us to analyze the impact of using WGAN (w/ f-AnoGAN) versus standard GAN architectures (w/ AnoGAN and MADGAN). The closest model to our WBHT approach for BH anomaly detection is obtained in f-AnoGAN model due to its improvement in training stability, unlike other GAN baseline models.  However, its performance is slightly lower than that of WBHT, likely due to its lack of structured spatial-temporal awareness. Overall, WBHT successfully integrates WGAN for stable training, LSTM-based encoding for sequential learning, and Multi-Head Attention for fine-grained feature extraction, allowing it to outperform all baseline models. For enhanced visual interpretability, Fig. \ref{fig:comparative} provides a comparative performance evaluation in terms of F1 and DR. The results demonstrate that our WBHT model consistently outperforms state-of-the-art alternatives, achieving the best F1 and DR scores. This confirms the effectiveness of our proposed approach in BH detection tasks, surpassing existing methodologies in terms of both accuracy and robustness.

\section{Conclusion} \label{sec:conc}

In this study, we proposed the WBHT framewotk for BH anomaly detection in communication networks using time series tabular data. Our approach integrates generative modeling, sequential learning, and attention mechanisms, building on WGAN architecture to enhance the model's stability and performance. %By incorporating a novel encoder-decoder framework, we achieved effective anomaly detection with minimal false positives and negatives. 

%The experimental results demonstrate the superior performance of WBHT compared to state-of-the-art models. 
Through a series of evaluations, we showed that WBHT outperforms traditional GAN-based models, Transformer-based architectures, and other advanced generative methods in BH detection tasks. Specifically, WBHT's ability to leverage LSTMMultiHead as the encoder and ConvLSTM as the generator enabled the model to effectively capture both spatial and temporal dependencies, making it highly suitable for real-time network anomaly detection. Our findings confirm that the combination of WGAN’s stability, the encoder’s temporal learning capabilities, and the generator’s spatial feature extraction significantly enhances the accuracy and robustness of BH detection. 

%In conclusion, WBHT represents a powerful tool for identifying sophisticated BH anomalies in network traffic, providing a promising solution for improving the security and reliability of communication networks. Future work will focus on ...
%\vspace{-8pt}

\section*{Acknowledgements}
This research is supported by the Scientific and Technological Research Council of Turkey (TUBITAK) 1515 Frontier R\&D Laboratories Support Program for BTS Advanced AI Hub: BTS Autonomous Networks and Data Innovation Lab. project number 5239903, TUBITAK 1501 project number 3220892, and the ITU Scientific Research Projects Fund under grant numbers MÇAP-2022-43823 and YESAP-2024-45920.

\bibliographystyle{IEEEtran}
\bibliography{main.bib}

\end{document}